\documentclass[letterpaper]{article}

\usepackage{authblk}
\title{An Introduction of mini-AlphaStar}
\author[1]{Ruo-Ze Liu\footnote{liuruoze@163.com}}
\author[1]{Wenhai Wang}
\author[1]{Yanjie Shen}
\author[1]{Zhiqi Li}
\author[1]{Yang Yu}
\author[1]{Tong Lu}
\affil[1]{Nanjing University, Nanjing, China}

\date{}
\usepackage{longtable}
\usepackage{booktabs}
\usepackage{graphicx}
\usepackage{float}
\usepackage{subfig}

\usepackage{amsmath}
\usepackage{algorithmic}
\usepackage{algorithm}
\usepackage{array}
\usepackage{xcolor}
\usepackage{amssymb}
\usepackage{hyperref}

\begin{document}
\maketitle
\begin{abstract}
	StarCraft II (SC2) is a real-time strategy game in which players produce and control multiple units to fight against opponent's units. Due to its difficulties, such as huge state space, various action space, a long time horizon, and imperfect information, SC2 has been a research hotspot in reinforcement learning. Recently, an agent called AlphaStar (AS) has been proposed, which shows good performance, obtaining a high win-rates of 99.8\% against human players. We implemented a mini-scaled version of it called mini-AlphaStar (mAS) based on AS's paper and pseudocode. The difference between AS and mAS is that we substituted the hyper-parameters of AS with smaller ones for mini-scale training. Codes of mAS are all open-sourced\footnote{\url{https://github.com/liuruoze/mini-AlphaStar}} for future research.
\end{abstract}

\section{Introduction}
\textbf{StarCraft II} (SC2) is a new challenge for reinforcement learning (RL)~\cite{sutton1998introduction}. To tackle it, DeepMind has proposed a learning environment called SC2LE (short for StarCraft II learning environment)~\cite{vinyals2017sc2} for researchers to build its agents on the SC2. The SC2LE provides a \textit{human-level} API for interacting between the agents and SC2. E.g., the agent should first select the unit by the unit's coordinate position in the game screen, and then the unit can be given a command. This schema greatly improves the learning difficulties. The agent not only needs to learn which action to do but also needs to learn how to select. Several works of research~\cite{Sun2018tsbot,pang2019sc2,tang2018sc2} have tried on this setting. But they only test against the built-in AIs. This situation is changed due to the proposition of \textbf{AlphaStar}~\cite{alpha}.

AlphaStar (AS) has beaten 99.8\% human players on Battle.Net (the battle matching platform for SC2). Compared to previous methods, AS has three differences: (1) huge computing resources. AS uses 128,000 CPU cores and 384 TPUs for training while lasting 44 training days; (2) a big neural network. AS's model consists of 3 encoders, 1 core, and 6 heads. Each of them contains one or multi convolution neural networks, recurrent neural networks, and Transformers~\cite{attention}; (3) usage of many new tricks, such as behavior cloning, pointer network, pFSP algorithm (a gaming theory method), and the scatter way in feature maps. These all contributed to AS's good performance. AS's success is not only an advance of technology but also an artwork of engineering.

However, AS is still not a perfect method to solve SC2 due to the following reasons. First, AS doesn't use the \textit{human-level} API in SC2LE (also called \textit{feature-level} API). AS used is a \textit{raw-level} API arising in the 3rd version of SC2LE's python interface, PySC2. The learning difficulties between the two are quite different. The exploring space for \textit{raw-level} API is indeed much smaller than \textit{human-level}. Thus reinforcement learning is easier. Second, the AS exploits human data in three ways. The overmuch usage of human data makes the learned agent more likely to imitate human behaviors instead of exploring new strategies. Third, the combat settings between AS and humans are not so fair (such as differences of EPM, camera moving operations, and control accuracy), making humans at a disadvantage in fighting against it, which reduced the persuasive results. For more information, we recommend the readers refer to another more detailed paper.

In this paper, we present our re-implementation of AS, called mini-AlphaStar (short as mAS). The design and implementation details are referenced by AS's Nature paper~\cite{alpha} and the supplementary details of that paper. We keep the structure and method of mAS similar to AS's but make many hyper-parameters much smaller ones (making it can be trained in a common commercial machine). We test the supervised learning code and reinforcement learning code of mAS and find it can be running as expected (however, the final performance of the training is not ideal). 

Section 2 presents the background knowledge of this paper. Section 3 introduces the deep neural network structure of mAS, the supervised learning and reinforcement learning of mAS, and the multi-agent mechanism. Section 4 gives the settings of mAS. Section 5 concludes the paper.
 
\section{Background}
Reinforcement learning is a sub-area of machine learning, which aims to obtain smart agents through training them within a specific environment to maximize the cumulative reward value. A popular research field in reinforcement learning is games. Initially, this method is used to solve simple games such as Black Jack and Atari games. With the rapid development of this field, attention has shifted to more complex games, as well as changes in research objects, from single-agent to multi-agent. As a typical multi-agent game, SC2 becomes a hot spot. 

Reinforcement learning has made some progress in the research of StarCraft. Oriol Vinyals et al.\cite{challenge} introduced the SC2  game environment, describing the characteristics and research challenges in this problem. Julian Togelius et al.\cite{map} devised a representation of StarCraft maps suitable for evolutionary search. Nicolas Usunier et al.\cite{episodic} applied episodic exploration for deep deterministic policies to StarCraft micromanagement tasks. Peng Sun et al.\cite{cheatlevel} presented their game agents that succeeded in defeating the cheating level built-in AI in the full game of SC2.

\section{Methods}
\subsection{Overall Structure}
The whole method consists of four main parts: Deep Neural Networks (DNN) architecture; Supervised Learning (SL) process; Reinforcement Learning (RL) process; and Multi-Agent (MA) League. 

DNN part is the base model for SL and RL. The agent was firstly trained by supervised learning using game replays of expert human players to obtain a base agent can imitate human players, which is implemented in the SL part. Then, the pre-trained agent will be trained through a series of self-play matches, using the reinforcement learning method to continuously adjust its policy network, which is implemented in the RL part. To improve the performance of reinforcement learning, a multi-agent league is used for collecting different trained agents and arranging matches between agents according to win-loss probability. The league also defines different kinds of agents with specific training targets, improving the training process by their cooperation. These selection algorithms for choosing which agent to be the opponents are implemented in the MA part.

\subsection{Neural Network Architecture}
The DNN architecture contains 3 encoder parts, 1 core part, and 6 head parts. The 3 encoders are entity encoder, spatial encoder, and scalar encoder. The entity encoder is used for processing the states of each entity to be one single tensor for further processing. The entity can be a unit or a building. The spatial encoder is used for processing the feature maps of the game screen or minimap. E.g., the height map or the visible map are processed in this encoder. The scalar encoder is for processing the statistical information of the game, e.g., the current minerals or supply of the game. The tensors output from the 3 three encodes are concatenated to input to the core part. The core part is a big LSTM which is to process the history information. LSTM accepts the input from the current state information from 3 encoders and the historical information from the hidden state. The output of the core part is going to the first head of the 6 heads, which is the action-type head. The action-type head decides which action type the agent will choose now. And its output is also carried on to the next head, which is the queue head. The output of the queue head decides the action whether should be queued. The output is also carried on to the next head. These transfers will be repeated. The next heads are delay head, selected units head, target unit head, and target location head respectively. The delay head decides how many frames the action will be delayed. The selected unit head decides which units will be put these actions. The target unit head and target location head decide which target will be chosen. The target will be an enemy unit or one position, so only one output of the two heads will be selected.

For RL training, the game process is pushed forward in frames. In each frame, the agent obtains the action by calling a step function. The step function accepts the observation and the previous hidden state, then outputs the action. The observation is called MsState. The three parts of MsState, the entity state, the spatial state, the scalar state, are pre-processed using different encoders respectively, and then be transferred to the core part. LSTM in the core part accepts the encoder input and generates the core output and the new hidden state. Then, the action type and an autoregressive embedding are got by passing the output of the core through a seris of processing. The action-type and the autoregressive embedding are used to generate other heads including the delay, the queue, the selected units, the target unit, and the target location respectively. Outputs from these heads form an ArgsAction. The ArgsAction will be transformed into a function call of PySC2 and will be applied in the environment.

\subsubsection{Encoder Structure}
\begin{figure*}[h!]
	\begin{minipage}[t]{\linewidth}
		\centering
		\includegraphics[width=\textwidth]{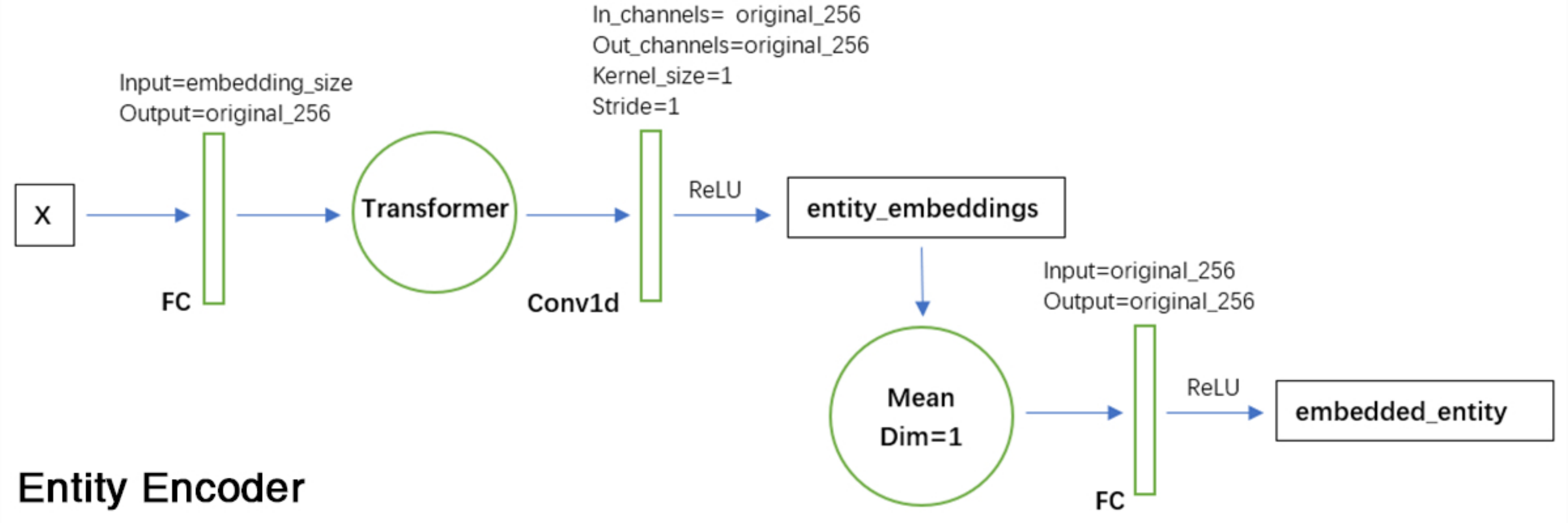}
		\caption{The entity encoder.}
		\label{fig:Encoders}
	\end{minipage}
\end{figure*}

We have different encoders for the three parts of MsState. In the entity encoder, the input is first embedded by a linear layer, then processed by a transformer. The output is passed through a ReLU and then a 2D convolution layer with ReLU to obtain entity-embedding. We also get embedded-entity by calculating the mean across all entities (the transformer output) and passing it through a linear layer with ReLU. Both two results are returned as the output of the encoder. The illustration of the entity encoder is shown in Fig.~\ref{fig:Encoders}.

In the spatial encoder, the input is first projected by a 2D convolution layer and a ReLU then downsampled through three 2D convolution layers, passed through several Resblocks to generate the map-skip, and finally through a linear layer with ReLU to generate the embedded-spatial. The results of the map-skip, the embedded-spatial are returned together as the output of the encoder.

In the scalar encoder, we first split the input scalar list into detailed elements. Each element is passed through the corresponding linear layer with ReLU and appended to a scalar list or context list. Then the elements in two lists are concatenated into a tensor and passed through a linear layer with ReLU respectively. The results, embedded-scalar-out, and scalar-context-out, are returned together as the output of the encoder.

\begin{figure*}[h!]
	\begin{minipage}[t]{\linewidth}
		\centering
		\includegraphics[width=0.8\textwidth]{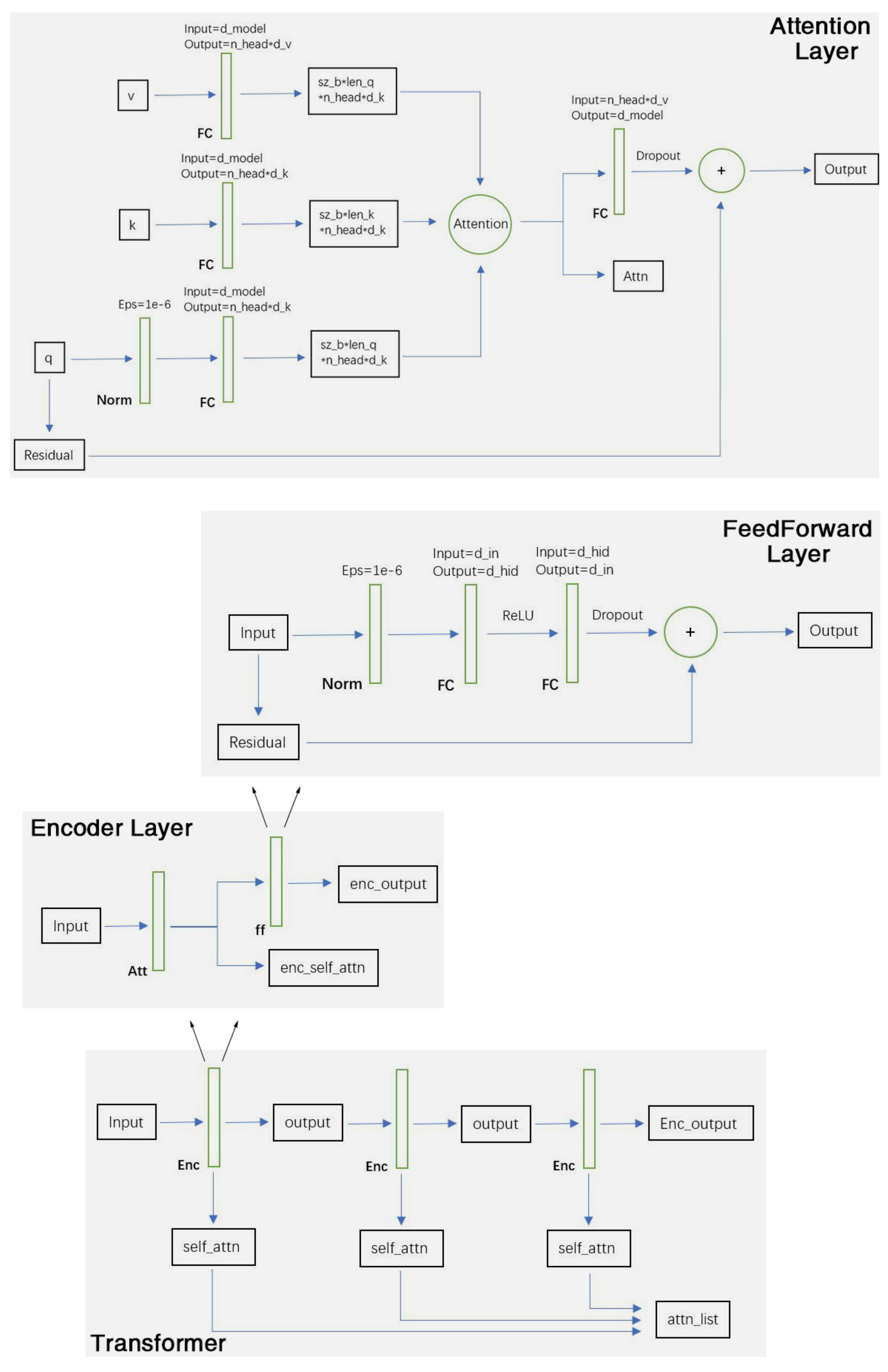}
		\caption{The transformer structure.}
		\label{fig:Transformer}
	\end{minipage}
\end{figure*}

For transformer structure in the entity encoder, it consists of three encoder layers, each of which contains two sublayers, an att (multi-head attention) layer, and an ff (position-wise feed-forward) layer. The att layer contains a residual network, which receives three inputs and generates an output value and an attention value. The ff layers receive the output of the att layer as input and also pass it through a residual network with two linear layers, one with ReLU and one with dropout. Finally, the result is returned together with the attention value. The output values are computed recursively through encoder layers, while attention values generated in each encoder layer are collected in a list. The illustration of the transformer structure is shown in Fig.~\ref{fig:Transformer}.

\subsubsection{Core Structure}
The core receives the previous hidden state and three embedded tensors (i.e., embedded-scalar, embedded-entity, and embedded-spatial). The embedded tensors are concatenated into one tensor. 
It is noted that before forwarding, the tensor shape is [batch\_seq\_size, embedding\_size],  we transform the shape to the [batch\_size, seq\_size, embedding\_size]. The LSTM model will accept the tensor and the previous hidden state, and output the new output and the new hidden state. 

\subsubsection{Head Structure}
We have six arguments in the action, which are action type, delay, queue, selected unit, target unit, and location. These components are generated in the following heads respectively. 

In the action-type head, we get LSTM output and scalar context as input. The LSTM output is first embedded through a linear layer with the shape original\_256, then through a series of Resblocks and ReLU. The result is then passed through a GLU gated by scalar context to generate action type logits. The action type is sampled from the probability obtained by applying a softmax on logits with temperature. Then we transfer the action type into one-hot and pass it through a linear layer with ReLU. The result and LSTM output are passed through GLU gated by scalar context respectively and added together to obtain autoregressive embedding. The logits, action-type, and autoregressive embedding are returned as output.

In delay head, we first decode the autoregressive embedding by passing it through two linear layers (each with size original\_256) with ReLU, then embed the result with a linear layer with shape last-delay. The delay value is sampled in a similar way as above. Then we transfer it to one-hot, pass it through two linear layers with ReLU, and finally project it to the shape same as autoregressive embedding. The result is added to the old autoregressive value and then returned as the new one, together with logits and delay value.

The way we deal with queue head is similar to that in delay head, except that the project queued value is applied a mask before added to autoregressive-embedding, which checks if queuing is possible for the chosen action type.

In the selected-units head, we first use a mask to determine which entity types can accept the action-type and generate a one-hot vector to represent it, then embed it through a linear layer with ReLU to obtain func-embed with shape original\_256. We also compute a mask representing which units can be selected, and a key corresponding to each entity passing the entity-embedding through a 1D convolution layer with original\_32 channels and kernel size 1. Then, we repeat the following procedure for selecting the max-limit of units: we first pass autoregressive-embedding through a linear layer with shape original\_256 and add it to func-embed, then pass the result through a linear layer with shape original\_32 and through ReLU, and feed it to LSTM (especially defined in this head) to generate query output and new hidden state. We multiply key and query, and then apply the mask computed above, obtaining the logits by dividing it by temperature. We compute entity-probs and sample entity-id as usual and append them to the logit list and unit list respectively. To update autoregressive-embedding, we transfer entity-id into one-hot and multiply it with key. We subtract the mean value of the one-hot from the result and project it through a linear layer, and finally add it to autoregressive-embedding. After the iteration, we concatenate the elements of logits and units respectively and return them together with autoregressive-embedding, each multiplied with a mask that represents whether action-type involves selecting units.

The process in the target-unit head is similar to what we do in the select-unit head, except that we do not update autoregressive-embedding since this head in one of the terminal heads. The other terminal head is the location head.

In the location head, we get autoregressive-embedding, action-type, and map-skip as input. We first concatenate autoregressive-embedding and map-skip along the channel dimension, then pass it through a ReLU and a 2D convolution layer with ReLU. The result is passed through a FiLM gated by autoregressive-embedding and then added with map skip. Afterward, it is unsampled by a series of transposed 2D convolution layers, then applied a mask decided by action-type to rule out the invalid locations, and we obtain the logits by dividing the result by temperature. Then we sample location-id using the probability of softmax value of logits and obtain location-out by calculating the coordinate from the index. Logits and location-out are both applied by a mask to check if the action-type involves targeting location and finally returned.

\subsection{Supervised Learning}

\subsubsection{Training data}
Training datasets are game replays from human players that were pre-processed into the feature label format. Features represent the game states of the player, while labels represent the actions taken by the player. The feature action pairs can be transformed into trajectories in the form of $[state, action, isfinal]$ for training use. A dataset object is constructed with replay data loaded by PyTorch. During learning, the supervised learner obtains training and testing data from it, and the predicted action value is generated by unrolling the state with network forwarding.

\subsubsection{Loss Function}
The label (action) to be learned contains six important parts, which are action type, delay, and queue. Therefore, the supervised training loss takes into consideration the corresponding loss, action type loss $t$, delay loss $d$, queue loss $q$, units loss $u$, target unit loss $tu$ if have, and target location loss $tl$ if have. Thus we have the loss function
\begin{equation} L=C(t_p,t_g)+C(d_p,d_g)+C(q_p,q_g)+C(u_p,u_g)+C(tu_p,tu_g)+C(tl_p,tl_g) \end{equation}
where $C$ denotes cross-entropy loss function, $p$ and $g$ are predict value and replay value respectively.

We additionally use a gradient clip of 0.5 to prevent gradient explosion and then apply updates to parameters with the Adam optimizer via gradient backward.

\subsection{Reinforcement Learning}
  
\subsubsection{State}
We maintain two states during training. One is the LSTM hidden state, which contains the information of all previous states and actions; the other is the MsState, which represents the current game state based on the player's observations, consisting of entity state, statistical state, and map state. The entity state contains information about the unit's status like unit position, health, attack/armor level, weapon cooldown; the statistical state represents some systemic properties like player/opponent race and upgrades, time, available actions; map state reveals the map information.

\subsubsection{Training process}
We use an actor-learner scheme for reinforcement learning. The actor thread and learner thread are created when the training starts. The actor thread continuously carries forward the game and keeps adding each other's trajectory into a list. When a specific number of trajectories are collected, they will be sent to the learner. When the number of trajectories received by the learner reaches batch-size, the learner extracts this batch and computes the loss function for it, and finally, network parameters are updated according to it. 

\subsubsection{Loss Function}
The loss function used in learner parameter update consists of four parts summing up.

The first part is actor-critic loss. It computes the pseudo-reward of actor trajectories and computes $TD(\lambda)$ with the baseline pseudo-reward associated with following the human strategy statistic z. It also computes the split vtrace pg-loss (policy-gradient), by calculating vtrace pg-loss respectively for ``action type'', ``delay'' and ``arguments'', which sums up pg-loss on all sets of logits of autoregressive actions. pg-loss is calculated using cross-entropy loss over logits and actions. The results above will be added up as a total actor-critic loss.

To compute the baseline, we first pre-process the observation of the player and the opponent and obtain each other's scalar-out value to form the action type together with the LSTM output. We pass it through a linear of original\_256 with ReLU, then through ResBlocks with $original\_256$ hidden units and layer normalization, and then through a ReLU and a linear with 1 hidden unit to get result $b$. Finally, we obtain the baseline value
\begin{equation}baseline=\frac{2}{\pi} arctan(\frac{\pi}{2} b)\end{equation}

The second part is split upgo-loss. We first remove the last timestep for trajectories and baselines, then calculate upgo-return from state values, step rewards, discount, and bootstrap values. Importance weight is computed by the difference between behavior log probability from cut trajectories and target log probability, then weighted advantage is obtained by multiplying the difference of upgo returns and state values with it. Finally, we compute the pg-loss on target logits and action type values obtained from trajectories, by multiplying it with the weighted advantage. The equation can be written as
\begin{equation}Loss=weight*C(logits_t,action)\end{equation}
$weight$ is computed as
\begin{equation}(upgo\_return-baseline)*exp(C(action,logits_t)-C(action,logits_b))\end{equation}
$upgo\_return$ is computed as
\begin{equation}lambda\_return(bootstraps,rewards,discounts,lambdas)\end{equation}
where $C$ denotes cross-entropy loss function, and $logits_t$ and $logits_b$ respectively denotes split\_target logits and behavior logits.

The third part computes the KL divergence between the actor trajectories to the supervised trained human policy, and the last part computes the entropy-loss for a set of policy logits.

\subsection{Multi-Agent League}
\subsubsection{League Architecture}
For self-play in reinforcement learning, we build a multi-agent league with three types of agents: main player (MP), main exploiter (ME), and league exploiters (LE). Main players are trained for improving themselves, while exploiters are trained to find the weakness of other agents. We also use a coordinator to maintain the payoff matrix and assigns new matches.

The payoff matrix is initialized when league training begins, containing the original agents. During the training process, outcomes of each match will be sent to the coordinator, updating the win-loss status between players in the matrix. 

For each training agent, checkpoints are set when its win rate to other historical players is higher than 0.7, or a specific number (like $4\times10^9$)of steps has passed. When an agent reaches the checkpoint, the current model of it will be added as a historical (i.e., past player) into the league.

\subsubsection{Game Matching}
The three types of agents have different training targets: main players train against all historical main players; main exploiters train against all players; league exploiters train against all historical players. Therefore, they have different mechanisms to get a match opponent

We use the priority fictitious self-play (PFSP) algorithm to randomly sample an opponent for a given player, by calculating the probabilities using a specific weighting function on win-rate, then randomly choose one according to normalized probability. Weight functions includes ``linear'', ``linear\_capped'', ``variance'' and ``squared'', and ``linear'' is set as default.

For main players, they have a 0.5 probability to directly sample a random historical player with squared weight; otherwise, they either try to find some rare players omitted or arbitrarily choose a main player and sample one of its historical with variance weight instead if it is too hard to beat.

For main exploiters, they first arbitrarily choose a main player. If the win rate against it is above 0.1, they will return it as the opponent; Otherwise, they will instead sample a random historical of it with variance weight.

For league exploiters, they directly sample a random historical player with linear\_capped weight.

\section{Setting}
In mini-AlphaStar, we substitute the hyper-parameters in AlphaStar, which includes races, batch size, network shape, etc., with smaller ones, in order to support mini-scale training. The detailed values are shown in the Table~\ref{tab:compare}, which compares the parameters between AlphaStar and mini-AlphaStar.
\begin{longtable}{|l|l|l|}
\hline
& AlphaStar & Mini-AlphaStar \\ \hline
Races & Protoss, Zerg, Terran & Protoss \\ \hline
batch\_size & 512/A\_s & 96/A\_s \\ \hline
sequence\_length & 64/A\_s & 64/A\_s \\ \hline
max\_entities & 64/A\_s & 32/A\_s \\ \hline
max\_selected & 512/A\_s & 384/A\_s \\ \hline
minimap\_size & 128 & 64 \\ \hline
embedding\_size & 3585 & 1543 \\ \hline
map\_channels & 18 & 18 \\ \hline
scalar\_encoder\_fc\_input & 1504 & 864 \\ \hline
scalar\_encoder\_fc\_input & 544 & 448 \\ \hline
scalar\_feature\_size & 7327 & 7327 \\ \hline
entity\_embedding\_size & 256 & 64 \\ \hline
lstm\_hidden\_dim & 384 & 128 \\ \hline
lstm\_layers & 3 & 1 \\ \hline
n\_resblocks & 16 & 4 \\ \hline
original\_1024 & 1024 & 256 \\ \hline
original\_512 & 512 & 128 \\ \hline
original\_256 & 256 & 64 \\ \hline
original\_128 & 128 & 48 \\ \hline
original\_64 & 64 & 32 \\ \hline
original\_32 & 32 & 16 \\ \hline
context\_size & 512 & 128 \\ \hline
location\_head\_max\_map\_channels & 128 & 32 \\ \hline
autoregressive\_embedding\_size & 1024 & 256 \\ \hline
baseline\_input\_size & 1152 & 1152 \\ \hline
league\_learner\_num & 12 & 4 \\ \hline
actorloop\_num & 16000 & 512 \\ \hline
\caption{Comparison of hyper-parameters. A\_s=Adjustable\_scale.}
\label{tab:compare}
\end{longtable}

\section{Conclusion}
AlphaStar is a program that achieves Grandmaster level in SC2 for the first time. We have implemented most of the components, with a few functions to be finished, including the z value generation. In the future, we will try to do more experiments on it for assessment and improvements.

\clearpage
\newpage

\section{Reference}
\vspace{-2.7em}

\renewcommand\refname{}

\bibliographystyle{unsrt}
\bibliography{ref}

\end{document}